\documentclass[journal]{IEEEtran}
\ifCLASSINFOpdf
\else
\fi
\usepackage{amsmath}
\usepackage[multi-part-units=single]{siunitx}
\usepackage{graphicx}
\usepackage{multirow}
\usepackage{bm}
\usepackage{cite}

\begin{document}
%
\title{Skin Lesion Diagnosis using Ensembles, Unscaled Multi-Crop Evaluation and Loss Weighting}
%
%
%

\author{Nils~Gessert$^{*ab}$,
		Thilo~Sentker$^{ac}$,
		Frederic~Madesta$^{ac}$,
		R\"udiger~Schmitz$^{ad}$,
		Helge~Kniep$^{ag}$,
		Ivo~Baltruschat$^{aef}$,
		Ren\'e~Werner$^{ac}$
        and~Alexander~Schlaefer$^{b}$
\thanks{$^*$Corresponding Author (e-mail: nils.gessert@tuhh.de)} 
\thanks{$^a$DAISYlab, Forschungszentrum Medizintechnik Hamburg, Germany}   
\thanks{$^b$Institute of Medical Technology, Hamburg University of Technology, 21073 Hamburg, Germany}
\thanks{$^c$Institute of Computational Neuroscience, University Medical Center Hamburg-Eppendorf, 20246 Hamburg, Germany}
\thanks{$^d$Department of Interdisciplinary Endoscopy and Institute of Anatomy and Experimental Morphology, University Medical Center Hamburg-Eppendorf, 20246 Hamburg, Germany}
\thanks{$^e$Section for Biomedical Imaging, University Medical Center Hamburg-Eppendorf, 20246 Hamburg, Germany}
\thanks{$^f$Institute for Biomedical Imaging, Hamburg University of Technology, 21073 Hamburg, Germany}
\thanks{$^g$Department of Diagnostic and Interventional Neuroradiology, University Medical Center Hamburg-Eppendorf, 20246 Hamburg, Germany}
}

%
%

\markboth{ISIC 2018 Challenge: Lesion Diagnosis}%
{Gessert \MakeLowercase{\textit{et al.}}: Skin Lesion Diagnosis using Ensembles, Unscaled Multi-Crop Evaluation and Loss Weighting}
%



\maketitle

\begin{abstract}

In this paper we present the methods of our submission to the ISIC 2018 challenge for skin lesion diagnosis (Task 3). The dataset consists of 10000 images with seven image-level classes to be distinguished by an automated algorithm. We employ an ensemble of convolutional neural networks for this task. In particular, we fine-tune pretrained state-of-the-art deep learning models such as Densenet, SENet and ResNeXt. We identify heavy class imbalance as a key problem for this challenge and consider multiple balancing approaches such as loss weighting and balanced batch sampling. Another important feature of our pipeline is the use of a vast amount of unscaled crops for evaluation. Last, we consider meta learning approaches for the final predictions. Our team placed second at the challenge while being the best approach using only publicly available data. 

\end{abstract}


%
\IEEEpeerreviewmaketitle

\section{Introduction}

Deep learning and in particular convolutional neural networks (CNNs) have become the standard approach for automated diagnosis based on medical images \cite{litjens2017survey}. For the problem of skin lesion diagnosis, a new dataset has recently been made available to the public \cite{Tschandl2018_HAM10000}. The dataset consists of 10000 dermoscopic images showing skin lesions which have been diagnosed based on expert consensus, serial imaging or histopathology. Using this dataset, the challenge "ISIC 2018: Skin Lesion Analysis Towards Melanoma Detection" has been proposed \cite{codella2018skin}. We participate in this challenge with an automated method that relies on multiple state-of-the-art CNNs, heavy data augmentation, loss weighting, extensive, unscaled cropping and meta learning. In the following, we describe the details of our approach. Our code is available to the public\footnote[1]{https://github.com/ngessert/isic2018}.

\section{Methods} \label{sec:methods}

\subsection{Evaluation Metric}

The key metric of this challenge is a weighted accuracy (WACC) across the seven classes. This is equivalent to the average recall or sensitivity. Hence, the metric is defined as:

\begin{equation}
	WACC = \frac{TP}{TP+FN}
\end{equation}

where $TP$ denotes true positive cases and $FN$ denotes false negative cases. These can be derived from a standard confusion matrix. We perform all preliminary evaluation and hyperparameter tuning with this metric.

\subsection{Dataset} \label{sec:dataset}

The baseline dataset is the HAM10000 dataset introduced by Tschandl et al. \cite{Tschandl2018_HAM10000}. In the following, we refer to this dataset as HAM. In addition, we used the public ISIC dataset which comprises roughly 13500 images. In the following, we refer to this dataset as ISIC. We checked all images for potential overlap between HAM and ISIC. 

The first obvious problem of these datasets is heavy class imbalance. Table~\ref{tab:datasets} shows the class distribution of HAM and ISIC. Therefore, we consider countering class imbalance as a key challenge to be addressed. Also, we have to assume that ISIC will not be that useful as it is even more imbalanced than HAM. For this reason, we only consider ISIC for few, high performing models in our final ensemble.

For internal validation, we split HAM into five sets with equal (imbalanced) class distribution for 5-fold cross-validation. All images were separated based on lesion affiliation, i.e., we made sure that images from the same lesion cannot occur both in a training and validation split. The information for this separation was provided by the organizers. We add the entire ISIC dataset to each training subset, if it is used. 


\begin{table}
	\caption{Class distribution of HAM and ISIC. MEL refers to melanoma, NV refers to melanocytic nevus, BCC refers to basal cell carcinoma, AKIEC refers to actinic keratosis, BKL refers to benign keratosis, DF refers to dermatofibroma and VASC refers to vascular lesion.}
	\label{tab:datasets}
	\centering
	\begin{tabular}{l l l l l l l l}
	 & MEL & NV & BCC & AKIEC & BKL & DF & VASC \\ \hline \\
	 HAM & $1113$ & $6705$ & $514$ & $327$ & $1099$ & $115$ & $142$ \\
	 ISIC & $1056$ & $11861$ & $72$ & $2$ & $477$ & $7$ & $0$ \\
	\end{tabular}
\end{table}

\subsection{Preprocessing} \label{sec:preproc}

For HAM, we kept the image size of $600\times 450$. Note, that histogram equalization was applied to selected images by the dataset publishers \cite{Tschandl2018_HAM10000}. For ISIC, we resized all images to the size of HAM using bicubic interpolation. 

During training, we applied online data augmentation to each image. First, we applied random cropping with a fixed size of $224\times224$. Note, that we did not use any scaling or aspect-ratio changes which are typically used \cite{Szegedy.2016b}. Then, we randomly flipped images along both dimensions with a probability of $p=0.5$. Furthermore, we distorted the images with random changes in brightness and saturation. Last, we subtracted the per-channel training set mean from the images. 

We tested random rotations, scaling, contrast, hue and aspect-ratio changes without an improvement in performance. 

\subsection{Models} \label{sec:models}

Overall, we rely on an ensemble for our final submission which has been successful in most challenges \cite{russakovsky2015imagenet}. In terms of model choice, we first assessed the value of using pretrained architectures. We found that fine-tuning a model trained on ImageNet performed signficantly better than training from scratch. Therefore, we chose to build our ensemble from pretrained models available to the public. We use the popular frameworks Tensorflow \cite{Abadi.2016} and PyTorch \cite{paszke2017automatic}. In particular, we use the Tensorflow Slim model library \cite{tensorflowslim} and the PyTorch pretrained models library \cite{pytorchmodels}.

The models to be included are selected based on 5-fold cross-validation performance. We tested the Inception \cite{Szegedy.2016b,Szegedy.2017} and ResNet \cite{He.2016,He.2016b} variants as a baseline first. These included InceptionV3, InceptionResNetV2, ResNet50-V1, ResNet50-V2 (post norm structure) and ResNet101-V2 (post norm structure). Next, we considered more recent architectures which included PolyNet \cite{zhang2017polynet}, ResNeXt \cite{xie2017aggregated}, Densenet \cite{huang2017densely}, SENets \cite{hu2017squeeze} and DualPathNet \cite{chen2017dual}. Compared to the baseline models, the more recent architectures performed better which is why we quickly excluded the baseline.

Thus, the models that are included in the final ensemble are variants of Densenet, ResNeXt, PolyNet and SENets. For most of our hyperparameter searches we used Densenet121 and assumed that the choices translate well to the other architectures.

\subsection{Training} \label{sec:training}

We found the training strategy to be crucial as well. We identified the most relevant hyperparameters to be the starting learning rate, the learning rate schedule and the choice of potential early stopping. 

Especially the latter is important in terms of model training for the final submission. Usually, the final submission model should be trained on all available data with a fixed learning rate schedule. During cross-validation, we performed early stopping which means that we saved a checkpoint of the model when the best WACC was achieved throughout the entire training process. We compared this to the last checkpoint being saved and noticed a significant difference in the WACC metric. In general, this implies that our chosen learning rate schedule is not optimal. However, we found it difficult to obtain the optimal learning rate schedule and the difference between the best and last model remained large. We did not observe this difference in such strength for normal accuracy or the area-under-curve (AUC) metric. We suspect that the WACC is very sensitive to changes of the classes with a small number of examples. Overall, this implies that our final model for submission could be trained using a validation set with early stopping instead of training a model on the entire available dataset. As this leads to suboptimal exploitation of the available data, we included both models from cross-validation with early stopping (5 models) and models trained on the entire dataset (1 model). The fully trained models are weighted $5$ times higher than individual CV models in order to achieve a reasonable balance.

The other hyperparameter choices are more straight forward. We tested Adam \cite{Kingma.2014}, Nadam \cite{dozat2016incorporating} (Adam + Nesterov momentum) and RMSprop in terms of optimizers and noticed only slight difference in performance with Adam generally performing best. We chose Adam for all models. In terms of learning rate schedule, we follow a stepwise approach. We chose a starting learning rate of $l_r = 0.0005$ and started reducing it with a factor of $\lambda = 0.2$ after $50$ epochs. Then, we continued reducing it with $\lambda$ every $25$ epochs. We stopped optimization after $125$ epochs. In between, we saved the model with the best WACC on the validation set, based on evaluation every $5$ epochs. We used a batch size of $40$.

For the loss function we used a standard cross-entropy loss as a basis:

\begin{equation}
	\mathcal{L} = - \sum_{c=1}^{C} p_c \log{\hat{p}_c}
\end{equation}

where $p$ is the ground-truth label, $\hat{p}$ is the softmax-normalized model output and $C$ the number of classes. As the seven classes in the dataset are highly imbalanced we considered different balancing approaches. In particular, we explored different ways of loss balancing and also sampling balanced batches. For loss balancing, we considered multiplying each classes' loss by its inverse normalized frequency, i.e., the weighting term $w_i$ is defined as

\begin{equation}
	w_i = \frac{N}{n_i}
\end{equation}

where $N$ is the total number of samples and $n_i$ is the number of samples for class $i$. This method puts a very strong weight on the highly underrepresented classes DF and VASC. Moreover, we considered a less extreme approach with

\begin{equation}
	w_i = \frac{N}{cn_i}
\end{equation}

\begin{figure*}[!ht]
\centering
\includegraphics[width=1\textwidth]{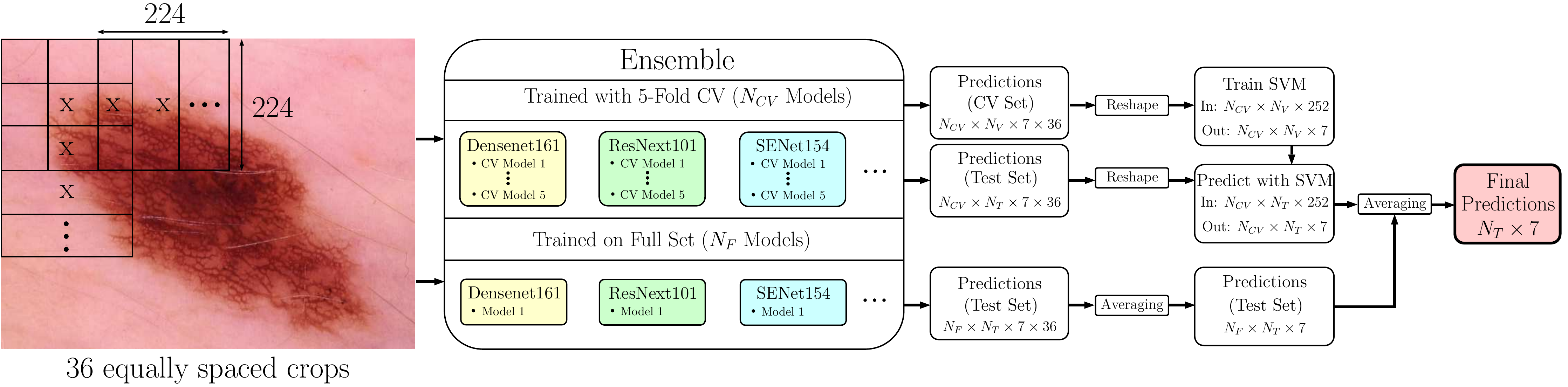}
\caption{Our evaluation strategy for the generation of the final predictions. $N_F$ denotes the number of models that were trained on the full training set. $N_{CV}$ denotes the number of models from 5-Fold CV training. $N_V$ is the number examples in the validation set and $N_T$ is the number of examples in the test set. }
\label{fig:eval}
\end{figure*}

where $c$ denotes the total number of classes. Last, we considered sampling balanced batches by oversampling the underrepresented classes during training. While all approaches improved results over no balancing at all, the differences were minor. We used the first loss balancing approach for all cases as it performed best for most models. It should be noted that we adjusted this method for use with HAM and ISIC combined. As ISIC is even more imbalanced than HAM, this balancing strategy would lead to unreasonable high loss amplification for underrepresented classes. Therefore, we derived the weighting from HAM only, both for training with HAM only and with HAM and ISIC.

Besides class balancing, we also tried to incorporate the meta information on the method of diagnosis. The organizers provided the information whether each lesion was diagnosed by "single image expert consensus", "serial imaging showed no change", "confocal microscopy with consesus dermoscopy" or "histopathology". Assuming that, within a given class, the means of diagnosis relates to the diagnostic difficulty, we tried to incorporate this meta information into the loss by increasing the loss for more difficult cases. Although this approach showed slightly increased performance for some models it appeared to be inconsistent across models and we did not incorporate it into all models of our final ensemble. E.g., for our reference model Densenet121 performance got even worse.

All training is performed on NVIDIA GeForce GTX 1080 Ti graphics cards. As some of the larger models have large memory requirements due to their feature map sizes, the graphics cards' memory was insufficient for our standard batch size. For these cases, we scaled down the learning rate and batch size by the same factor.

\section{Evaluation}

Our evaluation strategy for the generation of the final predictions is shown in Figure~\ref{fig:eval}. We made use of extensive multi-crop evaluation. The crops are unscaled and of size $224\times224$ which is identical to the size chosen for training. We perform $36$ evaluations per model which results in $36$ predictions that need to be combined. For the models that do not have a validation set, we performed averaging across the $36$ predictions. For the CV models, we incorporated a meta learning step. We constructed a flattened feature vector out of the $36$ predictions and used the results from the validation set for training an SVM. Then, we predicted the final label of the test set based on the $36$ CNN predictions. For this last model, we considered both random forests and SVMs with different kernels. We found SVMs with an RBF kernel to work best. Note, that a similar meta learning strategy was also used by one of last year's challenge winners \cite{menegola2017recod}. 

As a last step, we combined the predictions from the CV models and the fully trained models by averaging over all models. Instead of averaging we also considered voting, i.e., we counted how many models predicted a certain class. We found that averaging generally performs slightly better.

During training, we kept evaluating on the validation set using 16 crops in order to keep the computational effort at reasonable levels. For evaluation, we also tested increased numbers of crops, however, after 36 crops the improvement was negligible.

We performed model selection for our final ensemble based on the 5-Fold CV performance with $36$ crop evaluation of each architecture. For this selection, we simply averaged the predictions from all crops for evaluation. Then, we searched for an optimal combination of our architectures by averaging the predictions of a subset. In theory, an exhaustive search over all possible architecture combinations could be performed. However, this search would lead to millions of possible combinations which takes a significant amount of time. Instead, we first ranked all architectures by their 5-Fold CV performance. Then, we considered combinations where the best $15$ architectures are included only. We noticed that this approach usually leads to an optimal combination which includes roughly $6$ to $7$ out of the available architectures.

\section{Results} \label{sec:results}

\begin{table}
	\caption{5-Fold CV results for selected models and training scenarios. MACC denotes the mean accuracy. AUC is the mean area-under-curve value. In general, our previously described hyperparameter selection is used. Only differences are explicitly pointed out. For results with SVM we performed additional 10-fold cross-validation on each validation split. For the final ensemble we only considered the WACC metric.}
	\label{tab:results}
	\centering
	\begin{tabular}{l l l l}
	 & MACC & MAUC & WACC \\ \hline \\
	 Densenet121 & $0.823$ & $0.967$ & $0.795$ \\
	 Densenet121 \textbf{with SVM} & $0.827$ & $-$ & $\bm{0.822}$ \\
	 Densenet121 with ISIC & $0.870$ & $0.974$ & $0.804$ \\
	 Densenet121 no pretraining & $0.678$ & $0.931$ & $0.694$ \\
	 Densenet121 16-crop eval. & $0.822$ & $0.966$ & $0.785$ \\
	 Densenet121 4-crop eval. & $0.737$ & $0.932$ & $0.662$ \\
	 Densenet121 no weighting & $0.883$ & $0.976$ & $0.757$ \\
	 Densenet121 batch balancing & $0.797$ & $0.965$ & $0.789$ \\	 
	 Densenet121 diagnosis weighting & $0.874$ & $0.971$ & $ 0.738$ \\
	 Densenet161 & $0.861$ & $0.976$ & $0.809$ \\
	 Densenet169 & $0.852$ & $0.971$ & $0.806$ \\
	 ResNet50 & $0.862$ & $0.971$ & $0.779$ \\
	 SE-ResNet50 & $0.829$ & $0.966$ & $0.790$ \\
	 SE-ResNet101 & $0.838$ & $0.969$ & $0.810$ \\
	 ResNeXt101 32x4d & $0.836$ & $0.970$ & $0.808$ \\
	 SE-ResNeXt50 & $0.834$ & $0.968$ & $0.797$ \\
	 SE-ResNeXt101 & $0.860$ & $0.971$ & $0.803$ \\
	 DualPathNet92 & $0.862$ & $0.972$ & $0.804$ \\
	 \textbf{SENet154} & $0.854$ & $0.974$ & $\bm{0.817}$ \\
	 PolyNet & $0.845$ & $0.970$ & $0.802$ \\
	 \textbf{Ensemble} & $-$ & $-$ & $\bm{0.851}$ \\
	\end{tabular}
\end{table}

We report some preliminary results for the key parts of our approach. The results are derived from 5-Fold CV as the official validation set is not supposed to provide an indication of the performance on the test set. Since we did not use a held-out test set, we do not follow the procedure shown in Figure~\ref{fig:eval} for the results reported in this section. Instead, we simply average the predictions from the 36 crops and use them as the final prediction on each fold. We summarize the mean accuracy, mean AUC and WACC in Table~\ref{tab:results} for important architecture variations and an ensemble. In terms of models, we found that SENet performed best as a single model. Moreover, a large ensemble performs better than any single model approach. Our final ensemble contains 54 models with the following architectures: SENet154, ResNeXt101 32x4d, Densenet201, Densenet161, Densenet169, SE-Resnet101, PolyNet.

\section{Discussion and Conlusion} \label{sec:discussion}

In this paper we propose an approach for automatic skin lesion diagnosis for the "ISIC 2018: Skin Lesion Analysis Towards Melanoma Detection" challenge. We use a large ensemble of state-of-the-art CNN models. One of our key choices was to use full-sized images with unscaled, smaller crops for training in combination with extensive unscaled multi-crop evaluation. We combine the crops both by simple averaging and a meta learning strategy. This allows us to capture detailed, high-resolution features while also taking global context into account. Moreover, the HAM dataset is very challenging as it is highly unbalanced in terms of classes and the evaluation metric treats all classes equally. As this imbalance represents the real-world case where most examined lesions are benign, this is an important issue to be addressed. Therefore, we considered several balancing approach where simple loss weighting with inverse, normalized class frequency performed best. We also considered incorporating the meta information on how difficult it was to diagnose the lesion. We used the information by weighting the loss additionally by factor for each type of diagnosis. However, we observed inconsistent results across models. This indicates that our way of using the knowledge is not optimal. Also, the assumption that more extensive evaluation equals cases that are harder to learn is likely oversimplified. Finally, we constructed a large ensemble whose models were selected based on 5-Fold CV performance for our final predictions. Regarding single model performance, it is notable, that more recent architectures outperformed older standard architectures. Considering that many researches still use plain ResNets or even VGG as a baseline we suggest that it is reasonable to move to more recent architecture proposed for the natural image domain (ImageNet, etc.). With the overall goal of providing the best diagnosis we see this as an important step.
For future work, our method could be refined with a more extensive, less intuition-driven hyperparameter search. Moreover, the combination of local features and global context could be incorporated into a single end-to-end trainable architecture.

\section*{Acknowledgements}

This work was partially funded by the Forschungszentrum Medizintechnik Hamburg (02fmthh2017). Also, we gratefully acknowledge the support of this research by the NVIDIA Corporation under the GPU Grant Program.

\ifCLASSOPTIONcaptionsoff
  \newpage
\fi



%


\bibliographystyle{IEEEtran} 
\bibliography{egbib}

\begin{thebibliography}{10}
\providecommand{\url}[1]{#1}
\csname url@samestyle\endcsname
\providecommand{\newblock}{\relax}
\providecommand{\bibinfo}[2]{#2}
\providecommand{\BIBentrySTDinterwordspacing}{\spaceskip=0pt\relax}
\providecommand{\BIBentryALTinterwordstretchfactor}{4}
\providecommand{\BIBentryALTinterwordspacing}{\spaceskip=\fontdimen2\font plus
\BIBentryALTinterwordstretchfactor\fontdimen3\font minus
  \fontdimen4\font\relax}
\providecommand{\BIBforeignlanguage}[2]{{%
\expandafter\ifx\csname l@#1\endcsname\relax
\typeout{** WARNING: IEEEtran.bst: No hyphenation pattern has been}%
\typeout{** loaded for the language `#1'. Using the pattern for}%
\typeout{** the default language instead.}%
\else
\language=\csname l@#1\endcsname
\fi
#2}}
\providecommand{\BIBdecl}{\relax}
\BIBdecl

\bibitem{litjens2017survey}
G.~Litjens, T.~Kooi, B.~E. Bejnordi, A.~A.~A. Setio, F.~Ciompi, M.~Ghafoorian,
  J.~A. van~der Laak, B.~van Ginneken, and C.~I. S{\'a}nchez, ``A survey on
  deep learning in medical image analysis,'' \emph{Medical image analysis},
  vol.~42, pp. 60--88, 2017.

\bibitem{Tschandl2018_HAM10000}
P.~Tschandl, C.~Rosendahl, and H.~Kittler, ``The {HAM10000} dataset, a large
  collection of multi-source dermatoscopic images of common pigmented skin
  lesions,'' \emph{Sci. Data}, vol.~5, p. 180161, 2018.

\bibitem{codella2018skin}
N.~C. Codella, D.~Gutman, M.~E. Celebi, B.~Helba, M.~A. Marchetti, S.~W. Dusza,
  A.~Kalloo, K.~Liopyris, N.~Mishra, H.~Kittler \emph{et~al.}, ``Skin lesion
  analysis toward melanoma detection: A challenge at the 2017 international
  symposium on biomedical imaging (isbi), hosted by the international skin
  imaging collaboration (isic),'' in \emph{Biomedical Imaging (ISBI 2018), 2018
  IEEE 15th International Symposium on}.\hskip 1em plus 0.5em minus 0.4em\relax
  IEEE, 2018, pp. 168--172.

\bibitem{Szegedy.2016b}
C.~Szegedy, V.~Vanhoucke, S.~Ioffe, J.~Shlens, and Z.~Wojna, ``{Rethinking the
  inception architecture for computer vision},'' in \emph{{Proceedings of the
  IEEE Conference on Computer Vision and Pattern Recognition}}, 2016, pp.
  2818--2826.

\bibitem{russakovsky2015imagenet}
O.~Russakovsky, J.~Deng, H.~Su, J.~Krause, S.~Satheesh, S.~Ma, Z.~Huang,
  A.~Karpathy, A.~Khosla, M.~Bernstein \emph{et~al.}, ``Imagenet large scale
  visual recognition challenge,'' \emph{International Journal of Computer
  Vision}, vol. 115, no.~3, pp. 211--252, 2015.

\bibitem{Abadi.2016}
M.~Abadi, A.~Agarwal, P.~Barham, E.~Brevdo, Z.~Chen, C.~Citro, G.~S. Corrado,
  A.~Davis, J.~Dean, and M.~Devin, ``{Tensorflow: Large-scale machine learning
  on heterogeneous distributed systems},'' \emph{{arXiv preprint
  arXiv:1603.04467}}, 2016.

\bibitem{paszke2017automatic}
A.~Paszke, S.~Gross, S.~Chintala, G.~Chanan, E.~Yang, Z.~DeVito, Z.~Lin,
  A.~Desmaison, L.~Antiga, and A.~Lerer, ``Automatic differentiation in
  pytorch,'' in \emph{NIPS}, 2017.

\bibitem{tensorflowslim}
\BIBentryALTinterwordspacing
N.~Silberman and S.~Guadarrama, ``{TensorFlow-Slim image classification model
  library},'' 2016. [Online]. Available:
  \url{{https://github.com/tensorflow/models/tree/master/research/slim}}
\BIBentrySTDinterwordspacing

\bibitem{pytorchmodels}
\BIBentryALTinterwordspacing
``{Pretrained models for Pytorch},'' 2018. [Online]. Available:
  \url{{https://github.com/Cadene/pretrained-models.pytorch}}
\BIBentrySTDinterwordspacing

\bibitem{Szegedy.2017}
C.~Szegedy, S.~Ioffe, V.~Vanhoucke, and A.~A. Alemi, ``{Inception-v4,
  Inception-ResNet and the Impact of Residual Connections on Learning},'' in
  \emph{{AAAI}}, 2017, pp. 4278--4284.

\bibitem{He.2016}
K.~He, X.~Zhang, S.~Ren, and J.~Sun, ``{Deep residual learning for image
  recognition},'' in \emph{{Proceedings of the IEEE Conference on Computer
  Vision and Pattern Recognition}}, 2016, pp. 770--778.

\bibitem{He.2016b}
------, ``{Identity mappings in deep residual networks},'' in \emph{{ECCV}},
  2016, pp. 630--645.

\bibitem{zhang2017polynet}
X.~Zhang, Z.~Li, C.~C. Loy, and D.~Lin, ``Polynet: A pursuit of structural
  diversity in very deep networks,'' in \emph{Proceedings of the IEEE
  Conference on Computer Vision and Pattern Recognition}.\hskip 1em plus 0.5em
  minus 0.4em\relax IEEE, 2017, pp. 3900--3908.

\bibitem{xie2017aggregated}
S.~Xie, R.~Girshick, P.~Doll{\'a}r, Z.~Tu, and K.~He, ``Aggregated residual
  transformations for deep neural networks,'' in \emph{Proceedings of the IEEE
  Conference on Computer Vision and Pattern Recognition}.\hskip 1em plus 0.5em
  minus 0.4em\relax IEEE, 2017, pp. 5987--5995.

\bibitem{huang2017densely}
G.~Huang, Z.~Liu, K.~Q. Weinberger, and L.~van~der Maaten, ``Densely connected
  convolutional networks,'' in \emph{Proceedings of the IEEE Conference on
  Computer Vision and Pattern Recognition}, vol.~1, no.~2, 2017, p.~3.

\bibitem{hu2017squeeze}
J.~Hu, L.~Shen, and G.~Sun, ``Squeeze-and-excitation networks,'' in
  \emph{Proceedings of the IEEE Conference on Computer Vision and Pattern
  Recognition}, 2018.

\bibitem{chen2017dual}
Y.~Chen, J.~Li, H.~Xiao, X.~Jin, S.~Yan, and J.~Feng, ``Dual path networks,''
  in \emph{Advances in Neural Information Processing Systems}, 2017, pp.
  4467--4475.

\bibitem{Kingma.2014}
D.~Kingma and J.~Ba, ``{Adam: A method for stochastic optimization},'' in
  \emph{{ICLR}}, 2014.

\bibitem{dozat2016incorporating}
T.~Dozat, ``Incorporating nesterov momentum into adam,'' in \emph{ICLR}, 2016.

\bibitem{menegola2017recod}
A.~Menegola, J.~Tavares, M.~Fornaciali, L.~T. Li, S.~Avila, and E.~Valle,
  ``Recod titans at isic challenge 2017,'' in \emph{International Skin Imaging
  Collaboration (ISIC) 2017 Challenge at the International Symposium on
  Biomedical Imaging (ISBI)}, 2017.

\end{thebibliography}
%








\end{document}